\documentclass[letterpaper, 10 pt, conference]{ieeeconf}  

\usepackage{stix}
\usepackage{pifont}

\IEEEoverridecommandlockouts                              

\title{\LARGE \bf
FoMo: A Proposal for a Multi-Season Dataset for \\ Robot Navigation in For\^{e}t Montmorency
}

\author{Matěj Boxan$^{1}$, Alexander Krawciw$^{2}$, Effie Daum$^{1}$, Xinyuan Qiao$^{2}$,\\ Sven Lilge$^{2}$, Timothy D. Barfoot$^{2}$, and François Pomerleau$^{1}$
\thanks{$^{1}$Northern Robotics Laboratory, Université Laval, Quebec~City, Canada
		{\texttt{\footnotesize\{matej.boxan, francois.pomerleau\}@norlab.ulaval.ca}}
}
\thanks{$^{2}$University of Toronto Robotics Institute, Toronto, Canada
    {\texttt{\footnotesize\{alec.krawciw, samxinyuan.qiao, sven.lilge, tim.barfoot\}@robotics.utias.utoronto.ca}}
}%
\thanks{*This research was supported by the Natural Sciences and Engineering Research Council of Canada (NSERC) and the Fonds de recherche du Québec (FRQNT) through the grant 2023-NOVA-326877 HUNTER (Highlight the Unexpected with Navigation Through Extreme Regions).}
}

\usepackage[l2tabu,orthodox]{nag}

\usepackage[
    backend=bibtex8,
    style=ieee,
    sorting=none,
    natbib=true,
    doi=false,
    isbn=false,
    url=false,
    eprint=false,
    maxcitenames=1,
    mincitenames=1
]{biblatex}

\usepackage{amssymb,amsfonts,amsmath,amscd}

\usepackage{bm}

\newcommand{\bbm}{\begin{bmatrix}}
\newcommand{\ebm}{\end{bmatrix}}

\usepackage[pdftex,colorlinks]{hyperref}

\usepackage[french,english,noabbrev,nameinlink]{cleveref}

\usepackage[printonlyused]{acronym}

\usepackage{siunitx}
\usepackage[all]{nowidow}

\usepackage[dvipsnames]{xcolor}

\usepackage{lipsum}

\usepackage{xspace} 

\usepackage[pdftex]{graphicx}

\usepackage{epstopdf}

\usepackage{import}

\graphicspath{{./latexGoodPractices/}}

\usepackage{booktabs}

\usepackage{tabularx}
\usepackage{multirow, multicol}

\acrodef{UGV}[UGV]{Uncrewed Ground Vehicle}
\acrodef{IMU}[IMU]{Inertial Measurement Unit}
\acrodef{MEMS}[MEMS]{Micro-Electromechanical Systems}
\acrodef{GNSS}[GNSS]{Global Navigation Satellite System}
\acrodef{PTP}[PTP]{IEEE1588 Precision Time Protocol}
\acrodef{SLAM}[SLAM]{Simultaneous Localization and Mapping}
\acrodef{DOF}[DOF]{Degrees of Freedom}
\acrodef{NMEA}[NMEA]{National Marine Electronics Association}
\acrodef{PPK}[PPK]{Post Processed Kinematic}
\acrodef{RTK}[RTK]{Real Time Kinematic}
\acrodef{INS}[INS]{Inertial Navigation System}
\acrodef{GT}[GT]{Ground Truth}
\acrodef{ICP}[ICP]{Iterative Closest Point}
\acrodef{CAD}[CAD]{Computer Aided Design}
\acrodef{FoMo}[FoMo]{For\^{e}t Montmorency}

\newcommand{\stimes}{\vcenter{\hbox{$\scriptstyle\times$}}}

\usepackage{fancyhdr}
\fancypagestyle{withfooter}{
  
  \fancyhead[L]{}
  \fancyhead[R]{}
  \fancyfoot[C]{\footnotesize Accepted to the IEEE ICRA Workshop on Field Robotics 2024}
}
\begin{document}

\maketitle
\thispagestyle{withfooter}
\pagestyle{withfooter}

\begin{abstract}
In this paper, we propose the FoMo (Forêt Montmorency) dataset: a comprehensive, multi-season data collection.  
Located in the Montmorency Forest, Quebec, Canada, our dataset will capture a rich variety of sensory data over six distinct trajectories totaling 6~kilometers, repeated through different seasons to accumulate~42 kilometers of recorded data.
The boreal forest environment increases the diversity of datasets for mobile robot navigation.
This proposed dataset will feature a broad array of sensor modalities, including lidar, radar, and a navigation-grade \ac{IMU}, against the backdrop of challenging boreal forest conditions.
Notably, the \acs{FoMo} dataset will be distinguished by its inclusion of seasonal variations, such as changes in tree canopy and snow depth up to 2~meters, presenting new challenges for robot navigation algorithms.
Alongside, we will offer a centimeter-level accurate ground truth, obtained through \ac{PPK} \ac{GNSS} correction, facilitating precise evaluation of odometry and localization algorithms.
This work aims to spur advancements in autonomous navigation, enabling the development of robust algorithms capable of handling the dynamic, unstructured environments characteristic of boreal forests.
With a public odometry and localization leaderboard and a dedicated software suite, we invite the robotics community to engage with the \acs{FoMo} dataset by exploring new frontiers in robot navigation under extreme environmental variations.
We seek feedback from the community based on this proposal to make the dataset as useful as possible.
For further details and supplementary materials, please visit \url{https://norlab-ulaval.github.io/FoMo-website/}.
\end{abstract}

\section{INTRODUCTION}

Recent years have seen an unprecedented growth of mobile robots venturing into unstructured environments.
From caves and forests to deserts, this shift from urban centers to wild settings has been reflected in the development of new algorithms for \acf{SLAM}, semantic segmentation, traversability, and other tasks \cite{Ebadi2024,Baril2022,Schmid2022}.
Due to the costs of executing demanding experiments, especially in the wild, the community relies heavily on open-source datasets that serve as an essential benchmark tool, allowing the comparison of different approaches.
Moreover, as robotics research matures, new datasets provide new challenges.
Indeed, the past few years have seen releases of several high-quality datasets, focusing on outdoor environments with low texture \cite{Furgale2012}, poor lighting conditions \cite{Leung2017}, or seasonal changes \cite{Knights2023}. 

Motivated by these recent efforts, we focus our attention on boreal forests.
Despite being the largest land biome on Earth~\cite{Hayes2022}, boreal forests have been widely overlooked in the development of autonomous vehicles and remain, to the best of our knowledge, unexplored within the realm of a comprehensive robotics dataset.
While works similar to ours exist, they do not capture the desired range of sensors \cite{Ali2020}, \cite{Knights2023}, are mainly designed for tasks other than localization \cite{Tremblay2020}, or do not contain multi-season data.
It is worth noting that only a few works incorporate extensive online benchmarking tools and public leaderboards, both of which are essential prerequisites for the rigorous evaluation of various algorithms created by the community.

\begin{figure}[tb]
    \centering
    \includegraphics[width=\linewidth]{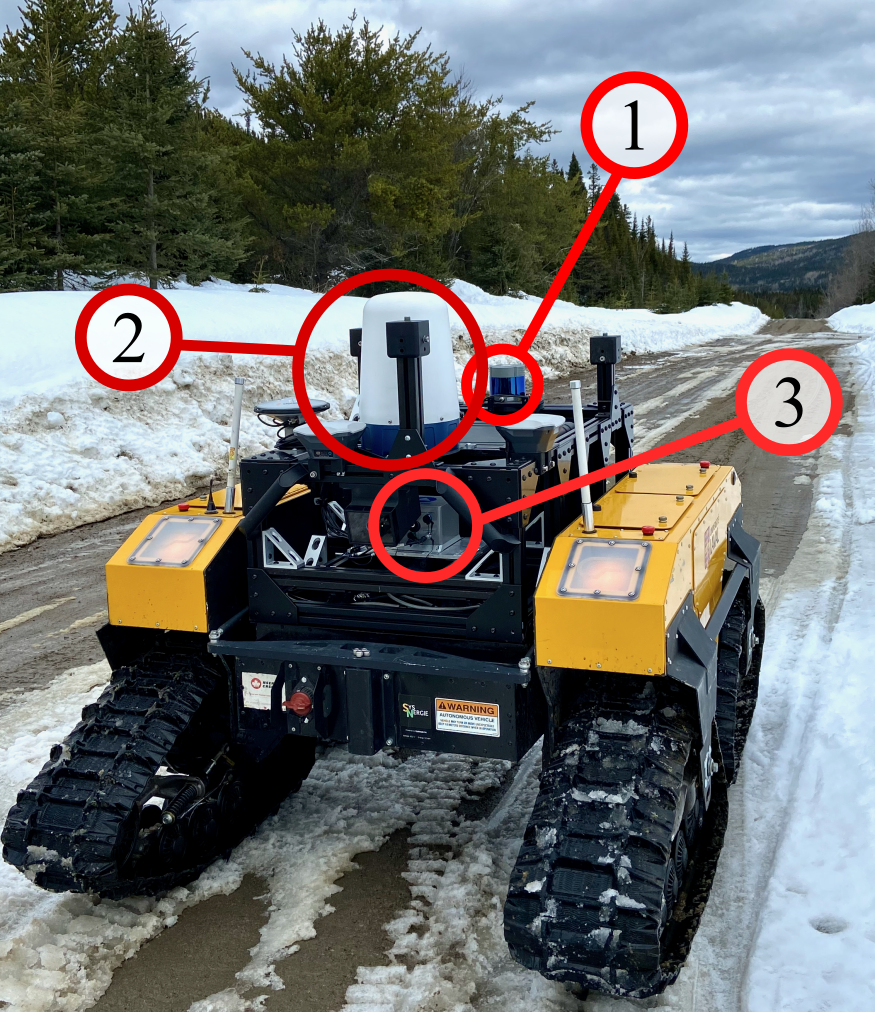}
    \caption{The data acquisition platform in the Montmorency Forest, Quebec, where the planned data recording will take place. 
    The platform can navigate through diverse terrain, including deep snow, supporting the multi-seasonal nature of the proposed dataset.
    Marked in red circles are the main sensor modalities: (1) lidar, (2) radar, (3) navigation-grade \ac{IMU}.
    }
    \label{fig:intro}
\end{figure}
In this article, we present the \acf{FoMo} dataset, our future work on a mobile robot navigation dataset recorded in a boreal forest.
We see this work as a proposal to the robotics community and welcome feedback on the presented sensor modalities, calibration procedures, and suggested trajectories.
The planned contributions of our work are as follows:
\begin{itemize}
    \item A rich, natural-environment dataset with six diverse trajectories spanning a combined length of \SI{6}{\kilo\meter} within a boreal forest.
    Due to regularly repeated experiments, the total length of the recorded data will reach \SI{42}{\kilo\meter}. 
    \item Seasonal changes, including tree canopy change and snow accumulation of up to \SI{2}{\meter}. 
    \item A wide range of sensing modalities, featuring a navigation-grade \acf{IMU} and a radar.
    \item
    Centimeter-accurate \acf{GT} trajectories, corrected with a \acf{PPK} \acf{GNSS} method.
    \item A public odometry and localization leaderboard, supported by a software suite to facilitate data manipulation.
\end{itemize}

\section{RELATED WORK}

\begin{table*}[ht!]\renewcommand{\arraystretch}{1.15}
	\centering
	\color{black}
	\caption{Overview of Different Navigation Datasets in Unstructured Environments.}
    \vspace*{-0.35cm}

    \parbox{15cm}{\centering
    The table is divided into five sections based on the environment.
    Datasets in individual sections are ordered by publication date.
    }
    \parbox{15cm}{\centering $\mdlgblkcircle$ - Sensor modality is included in the respective dataset. \hspace{0.5cm} $\mdwhtcircle$ - Sensor modality is not included in the respective dataset.}
    \vspace{0.35cm}
	\label{tab:sota_overview}
	\begin{tabular}{lp{1.75cm}p{1.75cm}cp{1.75cm}cccp{2cm}c}\toprule 
		 \textbf{Name \& References} & \centering \textbf{Environment} & \centering \textbf{Application} & \textbf{Camera} & \centering \textbf{Stereo Vision} & \textbf{Lidar} & \textbf{Radar} & \textbf{IMU} & \centering \textbf{GT} &\\
		\toprule
            Yamaha-CMU Dataset \cite{Maturana2018} & \centering Shrubland & \centering Segmentation & $\mdlgblkcircle$ & \centering $\mdwhtcircle$ & $\mdlgblkcircle$ & $\mdwhtcircle$ & $\mdlgblkcircle$ & \centering Pose (GPS), Segmentation&\\
            RELLIS-3D \cite{Jiang2021}  &  \centering Shrubland & \centering Navigation, Segmentation & $\mdlgblkcircle$ & \centering 800$\times$592 & $\mdlgblkcircle$ & $\mdwhtcircle$ & $\mdlgblkcircle$ & \centering Pose (D-GNSS), Segmentation&\\
            Tartandrive \cite{Triest2022} & \centering Shrubland, Woods & \centering Learning Dynamics  & $\mdlgblkcircle$ & \centering 1024$\times$512  & $\mdlgblkcircle$ & $\mdwhtcircle$ & $\mdlgblkcircle$ & \centering Pose (D-GNSS) &\\
            Goose Dataset \cite{Mortimer2023} & \centering Grassland, Woods & \centering Multi-Season Segmentation \& Perception& $\mdlgblkcircle$ & \centering $\mdwhtcircle$ & $\mdlgblkcircle$ & $\mdwhtcircle$ &  $\mdlgblkcircle$ &\centering Pose (D-GNSS), Segmentation &\\
            BotanicGarden \cite{Liu2024} &  \centering Grassland, Park & \centering Navigation, Mapping, Segmentation & $\mdlgblkcircle$ & \centering 1920$\times$1200  & $\mdlgblkcircle$ & $\mdwhtcircle$ & $\mdlgblkcircle$ & \centering Pose (Map-ICP), Map (Scanner), Segmentation&\\
          Oxford Offroad Radar Dataset \cite{Gadd2024} & \centering Grassland, Lake & \centering Navigation & \centering $\mdwhtcircle$ & \centering $\mdwhtcircle$ & \centering $\mdwhtcircle$ & \centering $\mdlgblkcircle$ & \centering $\mdlgblkcircle$ & \centering Pose (GPS) & \\
        \midrule
		Devon Island Dataset \cite{Furgale2012} & \centering Moon/Mars Analogue & \centering Navigation & $\mdlgblkcircle$ & \centering 1280$\times$960  & $\mdwhtcircle$ & $\mdwhtcircle$ & $\mdwhtcircle$ & \centering Pose (D-GPS) & \\
		Katwijk Beach Dataset \cite{Hewitt2018} & \centering Moon/Mars Analogue & \centering Navigation & $\mdlgblkcircle$ & \centering 1024$\times$768 \& 1280$\times$960  & $\mdlgblkcircle$ & $\mdwhtcircle$ & $\mdlgblkcircle$ & \centering Pose (GPS) &\\
		Mt. Etna Dataset \cite{Vayugundla2018} & \centering Moon/Mars Analogue & \centering Navigation & $\mdlgblkcircle$ & \centering 1292$\times$964 & $\mdwhtcircle$ & $\mdwhtcircle$ & $\mdlgblkcircle$ & \centering Pose (D-GPS) &\\
		MADMAX Dataset \cite{Meyer2021} & \centering Moon/Mars Analogue & \centering Navigation & $\mdlgblkcircle$ & \centering 1032$\times$772 \& 2064$\times$1544 & $\mdwhtcircle$ & $\mdwhtcircle$ & $\mdlgblkcircle$ & \centering Pose (GNSS) &\\
		\midrule
		Chilean Mine Dataset \cite{Leung2017} & \centering Mining Tunnel & \centering Navigation & $\mdlgblkcircle$ & \centering 1280$\times$960 & $\mdlgblkcircle$ & $\mdlgblkcircle$ & $\mdwhtcircle$ & \centering Pose (Markers \& Lidar)&\\
		SubT-Tunnel Dataset \cite{Rogers2020} & \centering Mining Tunnel & \centering Navigation, Mapping &$\mdlgblkcircle$  & \centering 2048$\times$1088 & $\mdlgblkcircle$ & $\mdwhtcircle$ & $\mdlgblkcircle$ & \centering Pose (SLAM), Map (Scanner) &\\
		\midrule
		Agricultural Dataset \cite{Chebrolu2017} & \centering Agricultural & \centering Navigation, Mapping, Plant Class. &  $\mdlgblkcircle$ & \centering 512$\times$424 (time-of-flight)& $\mdlgblkcircle$ & $\mdwhtcircle$ & $\mdwhtcircle$ & \centering Pose (GPS), Map (Scanner), Plant Class. Data  & \\
		Rosario Dataset \cite{Pire2019} & \centering Agricultural & \centering Navigation & $\mdlgblkcircle$ & \centering 672$\times$376 & $\mdwhtcircle$ & $\mdwhtcircle$ & $\mdlgblkcircle$ & \centering Pose (GPS) &\\
		\midrule
		{SFU} Mountain Dataset \cite{Bruce2015} & \centering Forest & \centering Navigation, Mapping & $\mdlgblkcircle$ & \centering 752$\times$480 & $\mdlgblkcircle$ & $\mdwhtcircle$ & $\mdlgblkcircle$ & \centering Pose (GPS), Location Correspondences &\\
		Freiburg Forest Dataset \cite{Valada2017} & \centering Forest & \centering Segmentation & $\mdlgblkcircle$ & \centering 1024$\times$768 & $\mdwhtcircle$ & $\mdwhtcircle$ & $\mdwhtcircle$ & \centering Segmentation &\\
		Canoe Dataset \cite{Miller2018} & \centering River, Forest & \centering Navigation & $\mdlgblkcircle$ & \centering 800$\times$600 & $\mdwhtcircle$ & $\mdwhtcircle$ & $\mdlgblkcircle$ & \centering Pose (GPS) &\\
		Montmorency Dataset \cite{Tremblay2020} & \centering Boreal Forest & \centering Navigation, Tree Class. & $\mdlgblkcircle$ & \centering $\mdwhtcircle$ & $\mdlgblkcircle$ & $\mdwhtcircle$ & $\mdlgblkcircle$ & \centering Pose (SLAM), Map (SLAM), Tree Class. Data &\\
		FinnForest Dataset \cite{Ali2020} & \centering Forest & \centering Multi-Season Navigation & $\mdlgblkcircle$ & \centering 1920$\times$1200 & $\mdwhtcircle$ & $\mdwhtcircle$ & $\mdlgblkcircle$ & \centering Pose (GNSS) &\\
		USVInland Dataset \cite{Cheng2021} & \centering Inland Waterways & \centering Navigation, Segmentation & $\mdlgblkcircle$ & \centering 640$\times$400 \& 1280$\times$800 & $\mdlgblkcircle$ & $\mdlgblkcircle$ & $\mdlgblkcircle$ & \centering Pose (GPS), Segmentation &\\
		FinnWoodlands Dataset \cite{Lagos2023} & \centering Forest & \centering Segmentation & $\mdlgblkcircle$ & \centering 1280$\times$720 & $\mdlgblkcircle$ & $\mdwhtcircle$ & $\mdwhtcircle$ & \centering Segmentation &\\
		Wild-Places \cite{Knights2023} & \centering Forest & \centering Long-Term Navigation, Mapping  & $\mdlgblkcircle$ & \centering $\mdwhtcircle$ & $\mdlgblkcircle$ & $\mdwhtcircle$ & $\mdlgblkcircle$ & \centering Pose (GPS), Map (SLAM)  &\\
		\toprule
		\textbf{\ac{FoMo} (Ours)} & \centering \textbf{Boreal Forest} & \centering \textbf{Multi-Season Navigation} & $\mdlgblkcircle$ & \centering  \textbf{1920$\times$1200} & $\mdlgblkcircle$ & $\mdlgblkcircle$ & $\mdlgblkcircle$ & \centering \textbf{Pose (GNSS) }&\\
		\toprule
	\end{tabular}
\end{table*}

In mobile robotics, datasets play an important role in evaluating and benchmarking trajectory and state estimation algorithms.
To date, a variety of such datasets have been proposed, featuring different robot manipulators, environments as well as sensor availability.
Early notable works include the MIT-DARPA \cite{Huang2010} and KITTI \cite{Geiger2012} datasets. 
Both feature a variety of sensors, as well as ground truth data via differential GNSS.
The environments featured in these early datasets are highly structured, mostly focusing on urban environments with static scenes while exhibiting idealistic weather and lighting conditions.
Consequently, evaluating algorithms on these datasets might not reflect their behavior in more challenging, less ideal environments and circumstances.

To overcome this shortcoming, additional datasets have been proposed to highlight and illustrate specific challenges and non-ideal features of real-world environments.
For instance, the KAIST Day/Night dataset \cite{Choi2018} features urban environments at different times throughout the day, leading to vastly different lighting conditions.
Both, the ComplexUrban \cite{Jeong2019} and UrbanLoco \cite{Wen2020} datasets highlight more challenging scenarios within urban environments, e.g., unreliable and sporadic GPS data, drifting \ac{IMU} measurements, complex building structures, as well as scenes with an abundance of dynamic objects.
Lastly, several datasets have been proposed that feature urban environments throughout different times of the year, highlighting the impact of weather and lighting in different seasons.
Prominent examples include the Oxford RobotCar \cite{Maddern2017}, the 4-Seasons \cite{Wenzel2021}, the Canadian Adverse Conditions \cite{Pitropov2021}, and the BOREAS Multi-Season \cite{Burnett2023} datasets.
While these efforts lead to datasets featuring more realistic scenarios, the aforementioned works are limited to highly structured, mostly urban environments.
However, in field robotics, robots are usually exposed to significantly unstructured environments, which introduce additional challenges to localization due to non-static scenes and few well-defined features.
Thus, existing datasets of structured environments are of limited use when evaluating field robotics algorithms and the development of specialized datasets featuring unstructured environments is warranted.

Although they have generally received much less attention, a variety of different datasets in unstructured environments have been published in the current literature.
\autoref{tab:sota_overview} shows an overview of relevant works, particularly highlighting the featured environment, the envisioned application, as well as the utilized sensors.
Datasets have been proposed for a variety of unstructured environments, ranging from off-road scenes \cite{Triest2022} to planetary surface analogues \cite{Furgale2012}, mining tunnels \cite{Leung2017}, agricultural farms \cite{Chebrolu2017} as well as natural environments such as gardens \cite{Liu2024} or waterways \cite{Cheng2021}.
In the following, we discuss the most relevant ones with respect to the work presented here, specifically focusing on forest environments.

The BotanicGarden dataset \cite{Liu2024}, targeted at navigation, mapping, and segmentation applications, features data collected in a natural garden environment using stereo vision, lidar, and \ac{IMU} data.
Both the Canoe \cite{Miller2018} and USVInland \cite{Cheng2021} datasets provide data collected along rivers and waterways in woodland and forest environments.
Both datasets include stereo camera images and \ac{IMU} data, while \citet{Cheng2021} further provide lidar and radar data.
Additional prominent forest navigation datasets are the {SFU} Mountain \cite{Bruce2015} and Montmorency \cite{Tremblay2020} datasets, featuring camera, lidar, and \ac{IMU} data.
\citet{Tremblay2020} additionally include stereo vision images and provide detailed tree classification labels.
The FinnWoodlands \cite{Lagos2023} and Freiburg Forest \cite{Valada2017} datasets mostly target semantic segmentation tasks.
Both datasets provide stereo vision data and detailed ground-truth segmentation labels, while \citet{Lagos2023} also feature lidar scans of the environment.
The Oxford Offroad Radar Dataset \cite{Gadd2024} is one of the few publicly available datasets including radar data in harsh weather conditions.
Similar to our proposal, the authors concentrate on repeated trajectories, in their case in the Scottish Highlands.

Typical applications of current forest datasets include autonomous navigation, mapping, and segmentation tasks.
However, only a subset of the studied navigation datasets feature data to evaluate long-term estimation and localization algorithms, which are crucial in changing unstructured environments.
Two such datasets are the FinnForest Dataset \cite{Ali2020} and Wild-Places \cite{Knights2023}.
While they both feature data collected in different seasons and over the course of several months, the included sensor readings are limited.
The FinnForest dataset lacks both radar and lidar readings of the environment.
Wild-Places, on the other hand, mainly focuses on lidar scans of the environment and similarly lacks radar.
Several cameras are recorded, but no matched stereo pairs are provided.
These shortcomings motivate the collection of the \ac{FoMo} dataset, which aims to collect long-term data in a boreal forest over several months, particularly focusing on capturing multi-seasonal changes and the gradual accumulation of snow.
A large variety of sensors are considered, including lidar, radar, stereo vision, and \ac{IMU} data.
We specifically envision that the use of radar, often neglected in other datasets, can be advantageous.
Unlike lidars and cameras, radar's performance does not degrade in harsh weather conditions, such as heavy snowfall.
Compared to lidars, it also offers a longer range of up to \SI{1}{\kilo\meter}, whereas lidars typically operate within lower hundreds of meters.

\section{THE \ac{FoMo} DATASET}
\begin{figure}[!htb]
    \centering
    \includegraphics[width=\linewidth]{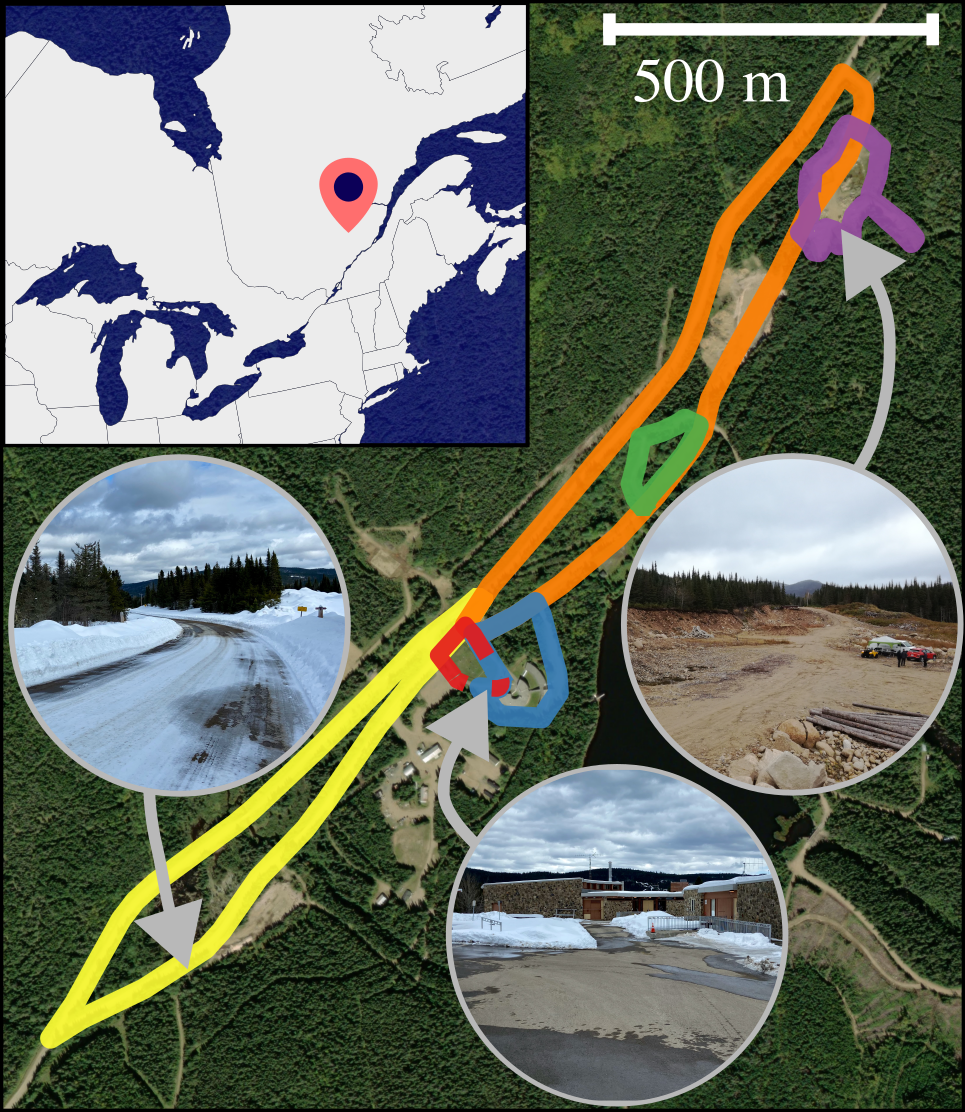}
    \caption{Satellite image of the Montmorency Forest, our data collection area, located in Quebec, Canada.
    The six proposed trajectories, overlaying the satellite image, have a combined length of \SI{6}{km}.
    In circles, we highlight images from the data collection site.}
    \label{fig:data-collection}
\end{figure}
The data will be gathered in the Montmorency Forest located \SI{70}{\km} north of Quebec City, Canada at a latitude of~\ang{47;19;15}N and a longitude of~\ang{70;9;0}W.
The site receives about \SI{900}{\mm} of rainfall and \SI{600}{cm} of snowfall each year.
The snow accumulation height reaches \SI{200}{\cm} in open areas and \SI{100}{cm} under tree canopy before the onset of snowmelt, which typically occurs at the end of April \cite{Hess2023}.
The proposed dataset will capture the seasonal changes in a typical example of boreal forests with dense coniferous vegetation, as well as deciduous species, including mostly birches and beeches.
We specifically target the changes in the tree canopy, as well as the accumulation of snow.

The planned data recording will start in early fall and continue once per month through winter up to spring, leading to approximately seven recording sessions.
We propose six trajectories with varying length and difficulty, shown in \autoref{fig:data-collection}.
The Red and Blue routes are situated in a semi-urban environment near the forest research center.
While the Red route has a total length of about \SI{290}{\meter} and serves as an entry point to get familiar with the data, the Blue route, with a total length of \SI{540}{\meter}, passes through a nearby forest with high snow cover in winter.
Similarly, the Green route (\SI{350}{\meter} long) traverses through a dense boreal forest marked by limited \ac{GNSS} coverage.
The Magenta route enters a stone quarry and has a total length of \SI{730}{\meter}.
Finally, the Orange and Yellow routes have a path length of around \SI{2000}{\meter}.
They follow maintained forest roads with good \ac{GNSS} signal coverage and will be the primary benchmarks of the proposed dataset.
In total, the described routes cover \SI{6}{\kilo\meter}, which leads to \SI{42}{\kilo\meter} of total length of the dataset.
The weather-resistant acquisition platform and proposed sensor configuration are described in the following sections.

\subsection{Acquisition Platform}
The acquisition platform, displayed in \autoref{fig:intro}, is a Clearpath Robotics Warthog \ac{UGV}.
The vehicle is equipped with four CAMSO ATV T4S tracks instead of wheels for the proposed dataset.
The skid-steered mobile platform has a differential suspension system.
Sufficient power autonomy is achieved with 16 Lithium-Ion battery modules with a total capacity of \SI{7.8}{\kWh}.
A custom-built modular sensor frame, detailed in \autoref{fig:sensor-cad}, is attached to the robot's chassis.
Mounted sensors and their calibration are described in detail in the next sections.

\subsection{Sensor Setup}
\label{sec:sensor_setup}

The proposed dataset will contain data from different proprioceptive and exteroceptive sensors. 
A ZED~X Stereo camera with 1920$\stimes$1200 resolution and RoboSense RS-32 3D lidar are mounted in front of the robot.
A Navtech CIR-304H radar is mounted in the rear, with the Atlans-C navigation grade \ac{INS} underneath it.
The Atlans-C uses integrated \ac{RTK} data from the Septentrio global navigation satellite system (GNSS) receiver.
Offering multiple types of \ac{IMU} data, we also include an MTi-10 2A8G4 and a VectorNav VN100 \ac{MEMS} \acp{IMU}.
In the rear, a Basler ace2 with a fisheye lens operating at 1920$\stimes$1200 points downwards.
As the robotics community shows increasing interest in multi-modal sensor data, for example as a supplementary input to neural networks, we also include an Audio-Technica ATR4650-USB microphone with a bit rate of \SI{128}{\kilo\bit\per\second}, mounted underneath the rear camera, and two Adafruit DPS310 barometric pressure sensors.
Precise \ac{GNSS} position is acquired with four Emlid ReachRS+ receivers.
One receiver is static and serves as a reference station while the other three are mounted on the \ac{UGV}.
Detailed sensor specifications can be found in \autoref{tab:sensors}.

\begin{table}[htb]
    \centering
    \caption{Sensor Specifications.}
    \label{tab:sensors}
    \begin{tabular}{lcr}
    \toprule
        Sensor Type & Model & Recording Frequency \\ \midrule
        Lidar                                   & RoboSense RS-32                   & \SI{10}{\Hz} \\
        \ac{IMU}                                & MTi-10 2A8G4                      & \SI{100}{\Hz}  \\
        \ac{IMU}                                & VectorNav VN100                   & \SI{200}{\Hz} \\
        \ac{INS}                                & Atlans-C                          & \SI{200}{\Hz} \\
        GNSS                                    & 4 $\times$ Emlid ReachRS+         & \SI{5}{\Hz} \\
        Radar                                   & Navtech CIR-304H                  & \SI{5}{\Hz}\\
        Pressure sensor                         & 2 $\times$ Adafruit DPS310        & \SI{100}{\Hz} \\
        Stereo camera                           & ZED X                             & \SI{10}{\Hz} \\
        Camera                                  & Basler ace2 + Fisheye Lens        & \SI{10}{\Hz} \\
        Wheel encoders                          & 2 $\times$ Hall effect sensors    & \SI{20}{\Hz} \\
        Microphone                              & Audio-Technica ATR4650            & \SI{16}{\kilo\Hz} \\
    \bottomrule   
    \end{tabular}
\end{table}

\subsection{Time Synchronization}
The proposed data collection system operates in a distributed manner. 
For this reason, it is critical to synchronize the clocks on every device. 
To minimize this discrepancy, the \ac{PTP} is enabled on the system.
The computer is the \ac{PTP} Grandmaster device because it is directly connected to the \ac{GNSS} receiver and maintains the most accurate time using the satellites' atomic clock.
\ac{PTP} is capable of synchronizing clocks within a few microseconds \cite{Vizzo2023}. 
Using \ac{PTP} with the combination with an \ac{GNSS} receiver is the current standard, outside of hardware triggering. 
Hardware triggering is infeasible for this system due to the additional wiring and compatibility requirements.
\ac{PTP} may require a few minutes to synchronize every clock. 
As part of the operational procedure, the robot will not record any data for at least five minutes after \ac{GNSS} lock to allow every device to update to the received time.

\begin{figure}[h]
    \centering
    \includegraphics[width=\linewidth]{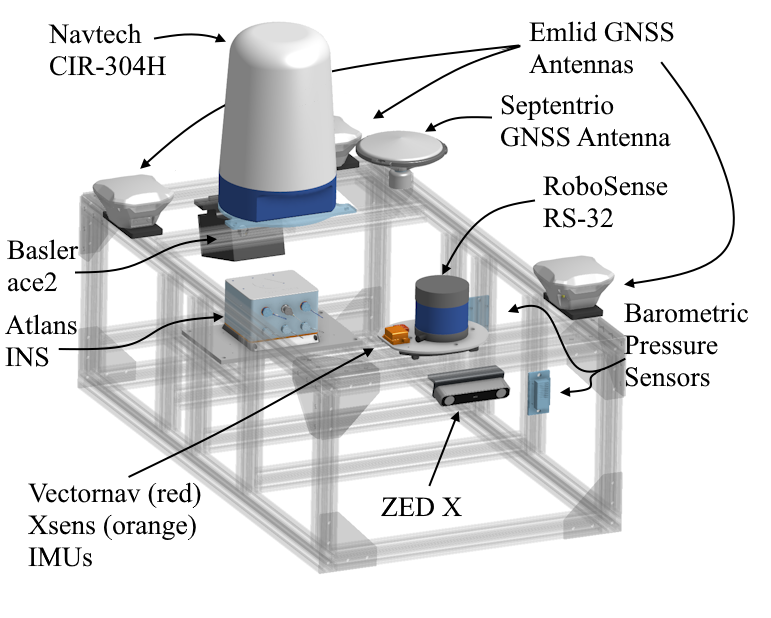}
    \caption{Rendering of the \acs{CAD} of the sensor frame, mounted on the \ac{UGV}.
    A lidar, a radar, two cameras, two barometric pressure sensors, three \acp{IMU} and four \ac{GNSS} antennas are attached to the frame.}
    \label{fig:sensor-cad}
\end{figure}

\subsection{Spatial Calibration}
The initial calibration estimate is determined from the mounting positions of the sensors in its \ac{CAD} model shown in \autoref{fig:sensor-cad}.
A calibration protocol for each sensor and sensor combination will be closely followed and executed before every deployment day.
The calibration data will be provided as part of every recorded sequence in a standard calibration format.

\subsubsection{Cameras}
The intrinsic characteristics of the ZED X are factory-calibrated through an extensive multi-step process.
The certified calibration procedure includes thermal measurements.
The Basler ace2 camera is calibrated with the ROS2 image\_pipeline library.\footnote{\url{https://github.com/ros-perception/image\_pipeline}}
Since the field of view of the Basler camera does not overlap with any other sensor, its position on the mounting system is taken from the \ac{CAD} model alone.

\subsubsection{Camera-\ac{IMU} Calibration}
The spatial calibration between the \acp{IMU} with respect to the ZED X camera is achieved with the Kalibr library.\footnote{\url{https://github.com/ethz-asl/kalibr}}
To allow for a wide range of movements needed for the calibration, the sensor frame is lifted with a crane and manipulated manually to excite all axes of the \acp{IMU}.
Our calibration target, also used for the intrinsic calibration of the Basler ace2 camera, is a chessboard \SI{765}{\milli\meter} wide and \SI{1000}{\milli\meter} tall.
 
\subsubsection{Camera-Lidar Calibration}
We employ the open-source calibration technique\footnote{\url{https://github.com/koide3/direct\_visual\_lidar\_calibration}} described by \citet{Koide2023} to find the transformation between the ZED X stereo camera and the RoboSense RS-32 lidar.
The method provides a target-less framework, relying instead on the structure of the environment.
The calibration will take place in the feature-rich space near the forest research center, depicted in the bottom circle in \autoref{fig:data-collection}.

\subsubsection{Lidar-Radar Calibration}
\begin{figure}[h]
    \centering
        \includegraphics[width=\linewidth, trim={100pt 200pt 0 100pt},clip]{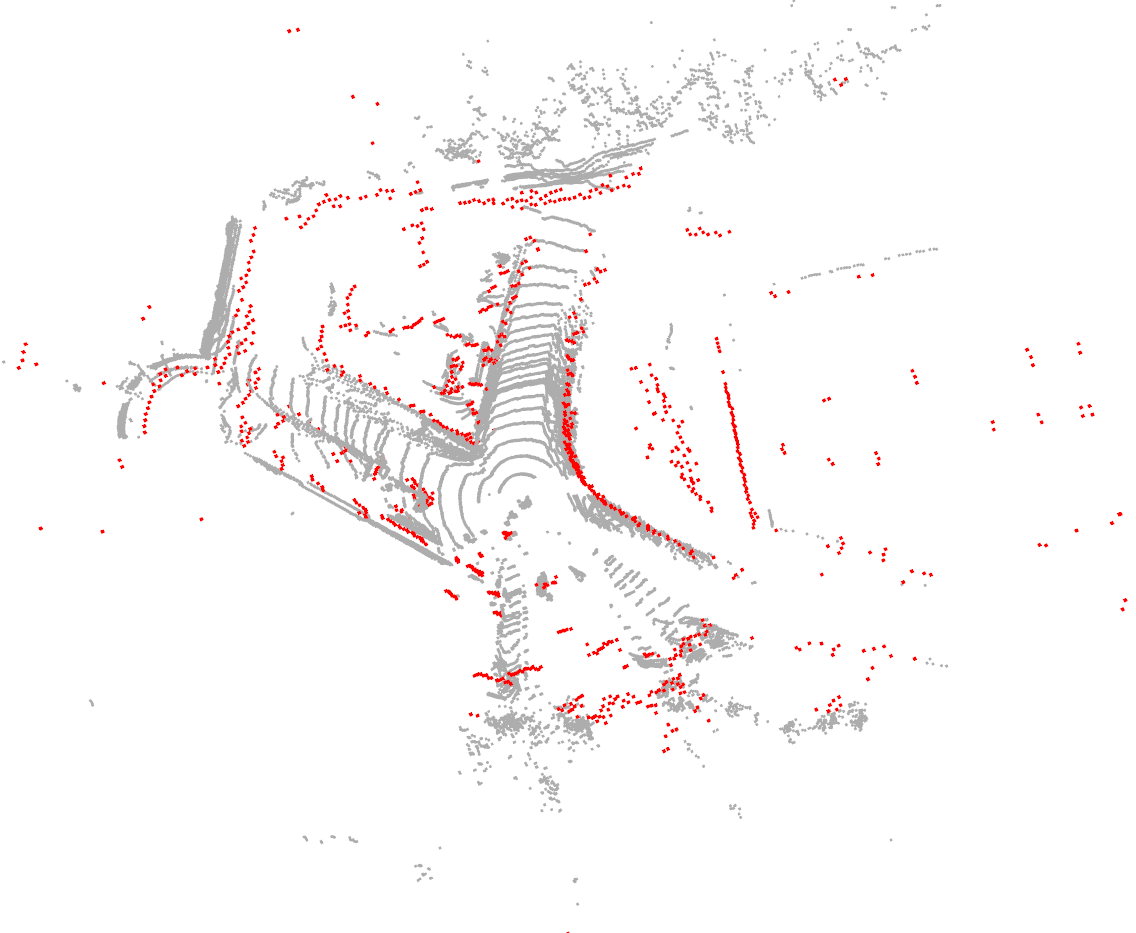}
    \caption{Lidar point cloud (grey) calibrated against extracted radar point cloud (red) using CACFAR~\cite{Burnett2022} as the extractor.}
    \label{fig:radar_lidar_calibration}
\end{figure}

The Navtech CIR-304H is a 2D spinning radar. For this reason, only the 3D position and 2D rotation will be estimated. 
Additionally, because the range resolution of the sensor is \SI{4.4}{\centi\meter}, the \ac{CAD} model will be used for the translation offset. 
The assembly tolerances are less than \SI{1}{\centi\meter}, which is unattainable from scan matching with these sensors. 
To estimate the heading offset between the two sensors, the \ac{ICP} algorithm is used.
The lidar point cloud is filtered by using the current estimate of the extrinsic calibration and the beam width to retain only points that could be measured by the radar. 
The radar's polar power image is converted into a point cloud using the CACFAR method \cite{Burnett2022}.
To ensure that the point clouds contain distinctive geometric features during calibration, this calibration step will occur in a structured environment near the deployment site.
\autoref{fig:radar_lidar_calibration} visualizes a lidar and extracted radar point clouds overlaid on top of each other.

\subsection{Data Collection}
In addition to the data captured by the onboard sensors, the dataset will contain comprehensive meteorological data from a weather station in the Montmorency Forest.
The data from the station include temperature and relative humidity measured with an HMP155A probe and barometric pressure obtained from a Vaisala CS106.
Both station have a refresh rate of \SI{1}{\minute}.
We will also include voltage and current readings from the batteries and both drive motors.
We recognize the importance of driver annotations of important events and conditions occurring during data recording.
These, along with driver IDs and other metadata, will be included with their timestamps for each dataset run.

\subsection{Ground Truth}
Given the planned scale of the dataset, it would be practically impossible to supplement it with \ac{GT} maps.
Indeed, seasonal changes would demand repeated scanning of the environment with a survey-grade stationery 3D laser before every deployment.
Moreover, heavy snowfall can cause significant changes to the environment, even on a span shorter than some of the proposed trajectories.
Since localization and mapping are interconnected problems, and considering the various combinations of sensors provided, we focus our efforts on providing precise \ac{GT} poses instead of a \ac{GT} map.

As noted in \autoref{sec:sensor_setup}, a static \ac{GNSS} antenna will be located in an open-sky area.
The coordinates of this antenna will be determined with a \ac{PPK} method in a geodesic coordinates system in the NAD83, MTM zone 7 projection.\footnote{\url{https://epsg.io/2949}}
Data from the three mobile antennas is recorded in the \ac{NMEA} format, enabling the computation of the full 6-\acl{DOF} \ac{GT} trajectories.
\ac{PPK} correction will be applied to the data, as the alternative \ac{RTK} method is less flexible due to its baseline restrictions and requirement for continuous signal.
On the contrary, the \ac{PPK} method does not require a strong connection between the three antennas on the \ac{UGV} and the base station, making it particularly suitable for boreal forests.
A geodetic pillar or a tripod fixed throughout the data recordings will ensure accurate and reproducible results~for~the~\ac{GT}.

\section{PROPOSED DATASET USAGE}
We envision the primary use of our dataset for 3D odometry and metric localization.
Such applications align with other autonomous driving datasets; however, we believe that the multi-seasonal conditions in a boreal forest will challenge established odometry and \ac{SLAM} pipelines.
The recorded data will also present a good opportunity for terrain-based power estimation.
Last but not least, the comprehensive recorded data might be useful in diverse tasks, such as semantic segmentation, terrain classification, traversability estimation, and multi-sensor fusion.

\section{CONCLUSION AND FUTURE WORK}
This paper proposes the \ac{FoMo} dataset for robot navigation in sub-arctic environments. By repeating the same routes through the fall-winter-spring freeze-thaw cycle, research into autonomy that is robust against significant environmental variation can advance.
The \ac{FoMo} dataset will contain about \SI{6}{km} of paths at each recording session and \SI{42}{km} in total. 
By pairing the sensor data from the \ac{UGV} with the weather data available at Montmorency Forest, researchers will be able to investigate the effects of ambient weather on the traversability of the terrain.
The \ac{PPK} GNSS ground truth poses will be used for quantitative evaluation of odometry and localization pipelines. 
We anticipate research that studies each sensor modality individually, as well as sensor fusion approaches.
The public leaderboard will be available on the dataset website after the final release. 

\printbibliography

\end{document}